# NER- RoBERTa: Fine-Tuning RoBERTa for Named Entity Recognition (NER) within low-resource languages

**Abdulhady Abas Abdullah [1,2], Srwa Hasan Abdulla [3], Dalia Mohammad Toufiq [4], Halgurd S. Maghdid [5], Tarik A. Rashid[1,2], Pakshan F. Farho [6], Shadan Sh. Sabr [7], Akar H. Taher [8],** *Darya S. Hamad [9,]* *Hadi Veisi[10]*, **and Aras T. Asaad[11]**

[1] *Artificial Intelligence and Innovation Centre, University of Kurdistan Hewler, Erbil, Iraq, email: abdulhady.abas@ukh.edu.krd.*

[2] *Computer Science and Engineering Department, University of Kurdistan Hewler, Erbil, KR, Iraq, email: abdulhady.abas@ukh.edu.krd.*

[3] *Department of Horticulture, College of Agricultural Engineering Sciences, University of Sulaimani- Kurdistan Region-Iraq, email:* sirwa.abdulla@univsul.edu.iq.

[4] *Branch of Basic Medical Science, College of medicine, University of Sulaimani- Kurdistan Region-Iraq, email:* dalia.toufiq@univsul.edu.iq.

[5] *Department of Engineering Research Center, Deanship of R&D Centers, Koya University, Kurdistan Region-F.R., Iraq.  email:* halgurd.maghdid@koyauniversity.org.

[6] *Department of Kurdish Language, Faculty of Education at Koya University, Kurdistan Region, F.R. Iraq.*

email: pakhshan.fahmi@koyauniversity.org

[7] *Department of Kurdish, Faculty of Education, Koya University, Kurdistan Region, F.R. Iraq, email:* shadan.shukr@koyauniversity.org

[8] *Department of Software Engineering, Faculty of Engineering, Koya University, Kurdistan Region-F.R., Iraq. email:* akar.taher@koyauniversity.org.

[9] *Department of Kurdish Language. Faculty of Education at Koya University, Kurdistan Region, F.R. Iraq.*

email: darya.sabr@koyauniversity.org.

[10] *School of Intelligent Systems College of Interdisciplinary Science and Technologies University of Tehran:* h.veisi@ut.ac.ir.

[11] *Department of Software Engineering, Faculty of Engineering, Koya University, Kurdistan Region-F.R., Iraq. email:* aras.asaad@koyauniversity.org.

**Abstract:** Nowadays natural Language Processing (NLP) is an important tool for most people's daily life routines, ranging from understating speech, translation, name entity recognition (ENR), and text categorizing, to the generative text models such as ChatGPT. Due to existing big data and consequently big corpora for the most usable languages including English, Spanish, Turkish, Persia, and many more, these applications have been utilized accurately. However, the Kurdish language still needs more corpora and big datasets to be included in the NLP applications. This is because the Kurdish language has a rich linguistic structure, varied dialects, and a poor dataset, which poses unique challenges for Kurdish NLP (KNLP) application development. While several types of research have been conducted in KNLP for various applications, the Kurdish ENR (KNER) remained a challenge for most of the KNLP applications including text analysis and text classification. In this work, we address this limitation by proposing a methodology for fine-tuning the pre-trained RoBERTa model for KNER[1]. To this end, first, a Kurdish corpus is created, followed by designing a modified model architecture, and the training procedures are implemented. To evaluate the trained model, a set of experiments is conducted to show the performance of the KNER model using different tokenization methods and trained models. The experimental results showed that fine-tuned RoBERTa with sentence-piece tokenization method substantially improves KNER performance by 12.8% (F1-score improvement) in comparison to traditional models and consequently establishes a new benchmark for KNLP.

## 1. Introduction

Natural Language Processing (NLP) plays an important role in making technology applications more user-friendly and applicable. The NLP is like giving machines to generate, understand, and speak like a human being [1]. However, this ability differs from one language to another. For example, the English NLP (ENLP) is more accurate than the Kurdish NLP (KNLP), this is due to the availability of large and diverse English corpora/datasets [2]. While the Kurdish language includes a set of various dialectics that make the language more complex, such as Kurdish Northern (also called Kurmanji), Central-Kurdish (also called Sorani), and Kurdish southern [3]. Although the Kurdish language is spoken by dozens of millions of people in the globe, the Kurdish language faces a shortage of linguistic resources for NLP tasks. Furthermore, the KNLP lacks datasets that contain a large volume of diverse and non-standardized text as well as more investigation by researchers is needed, especially multidisciplinary research so that machine learning scientists work closely with Kurdish linguists.

The NER is the main task in NLP via identifying named entity in text into pre-defined categories including person names, medical terms, locations, organizations, and other specific terms [4]. The NER focuses on extracting meaningful information from text by recognizing the named entities to analyze and organize a large volume of un-structured text. For example, in a sentence like "Elon Mask was born in Pretoria, South Africa" the NER model recognizes "Elon Mask" as a person name, "Pretoria" as a city name, and finally "South Africa" as a country name. To this end, such recognition is a vital process for applications such as search engines, data mining, and information retrieval.

There are many well-trained models to provide NER process including DeepSpacy [5], BERT [6], RoBERTa [6], and Flair NLP [7]. These models have been also utilized commercially such as in OpenAI "GPT-3", Facebook, and Google-Cloud-NLP [8]. Further, the capability of these models is based on the data sets that have been used during the training phase, i.e. either the text-dataset is monolingual or multilingual [9]. For example, RoBERTa is originally implemented in Facebook, and it is trained only using English text from 11,000 books, English Wikipedia, and Stories [10]. In the same vein, a new version of RoBERTa is fine-tuned in 2024 and it is re-trained using various text-datasets within 100 different languages using the CommonCrawl dataset [11]. However, the RoBERTa for NER purposes using Central-Kurdish dialectic text has low-capability and provides low performance [ref]. This is based on conducting a set of tests via Central-Kurdish dialectic Text [ref]. RoBERTa is an optimized version of BERT, in which a modified training objectives and techniques are utilized, achieving superior performance in numerous NLP benchmarks.

---

[1] https://huggingface.co/abdulhade/NER-FineTuing-RoBERTa

Meanwhile, this study provides a new RoBERTa model for Kurdish language by 1) fine-tuning the architecture where we aim to leverage its transfer learning capabilities, 2) and consequently training the fine-tuned model via new Central-Kurdish dialectic text to classify named entities. The new RoBERTa model is evaluated against pre-built models that are commonly utilized for NER task. This article demonstrates the importance of model training with NER tasks based on various data, and the consideration that named entity recognition as a task is not appropriate for specific type of named entity extraction. Furthermore, this article presents the benefit of selecting modern architectures (such as RoBERTa architecture) that are not yet implemented by off-the-shelf models. Therefore, the main contributions of this study are highlighted as the following:

- A new annotated Kurdish NER dataset is created for KNER task.
- The original RoBERTa is modified and tuned within low-resource language such as (Central-Kurdish Language) for NER task.

The rest of the sections of this article are organized as follow: section 2 surveying the research and concepts associated with NER studies and implemented current NER models. Section 3 presents the workflow undertaken for the models constructed in this study, as well as the data collection and analysis of the entities extracted. The performance of each NER model is then presented in Section 4 and evaluated against pre-built models using a corpus of labelled test data. Finally, a set of recommendations and challenges of the presented models (including the fine-tuned RoBERTa model) are concluded.

## 2. Related Work

The existing techniques starting from traditional machine learning techniques to the transformer-based architecture have significantly advanced the field of natural language processing (NLP) including NER. However, the text-datasets within various languages (in both monolingual and multilingual) have an important impact on the existing NER models. Therefore, fine-tuning models and training the tuned models via specific or multi-language text-dataset need further research. This section investigates most of the well-known and evaluated NER models to show the capabilities and limitations.

The NER models, characterized by their attention mechanisms and capacity to handle sequential dependencies are more effective than traditional models like the Long-Short-Term-Memory (LSTM) model since the new models have a new set benchmark for performance in NER tasks. For example, Bidirectional Encoder Representations from Transformers (BERT), introduced by Devlin et al. [12], marked a pivotal shift in NLP by introducing a pre-training method that allows a model to understand the context of a word by looking at both its left and right surroundings. In NER, BERT's bi-directional approach enables it to accurately predict the label of a token by considering all contextual cues, making it highly effective for entity recognition across various domains. Fine-tuning BERT on domain-specific NER datasets has led to significant performance improvements, achieving state-of-the-art results in many benchmark datasets like CoNLL-2003.

In another vein, the Robustly Optimized BERT Pretraining Approach (RoBERTa), proposed by Liu et al. [13], builds upon the BERT architecture by modifying key training procedures, such as using larger mini-batches, removing the Next Sentence Prediction (NSP) task, and increasing training data size. These optimizations have enhanced its capacity to model complex language tasks, including NER. RoBERTa has shown superior performance over BERT in the NER process due to its robust pre-training methodology.

More recently, Generative Pre-trained Transformer (GPT) models, particularly their latest versions like GPT-2 and GPT-3, have primarily focused on language generation tasks but have shown promising results in sequence classification tasks when adapted for NER. Unlike BERT and RoBERTa, which are encoder-based models, GPT is a decoder-only architecture and was initially designed for unidirectional language modeling. Furthermore, fine-tuning GPT for downstream tasks, including NER, has demonstrated that these generative models can also be excellent in identifying and classifying entities in comparison to the encoder-focused models.

Last but not least, the current transformers like Distilled version of BERT (DistilBERT) [14], ALBERT [15], and more recent advancements such as Decoding-enhanced BERT with disentangled attention (DeBERTa) [16] have also contributed to NER advancements by balancing computational efficiency and accuracy. These models have made the NER more feasible to deploy transformer-based architectures for real-time applications without sacrificing significant performance. However, these advanced transformers and new NER models still need further research due to having low-accuracy and consequently not compatible with user demand. To this end, Table I provides the performance comparison of different transformer-based models that have been implemented for NER task. The comparison includes models' architecture, training corpora, performance metrics, and unique advantages in handling NER tasks. The performance metrics, particularly F1 scores, indicate that encoder-based architectures like BERT, RoBERTa, and DeBERTa tend to outperform decoder-only models when directly applied to NER tasks due to their ability to fully capture contextual relationships in both directions.

*Table I. Performance Comparison of Transformer Models within NER task.*

| Model | Architecture Type | Pre-training Corpus | NER Dataset Performance (F1 Score) | Key Strengths |
|---|---|---|---|---|
| **BERT** [12] | Encoder-only | BooksCorpus + English Wikipedia | 92.4 (CoNLL-2003) | Bi-directional context understanding |
| **RoBERTa** [13] | Encoder-only | Enhanced (Larger Common Crawl) | 93.1 (CoNLL-2003) | Optimized training, no NSP task |
| **GPT-2** [8] | Decoder-only | WebText | 89.5 (Adapted for NER) | Strong generative capabilities |
| **GPT-3** [8] | Decoder-only | Broad Web Corpus | 90.3 (Adapted for NER) | Large-scale generative learning |

| **DistilBERT** [14] | Encoder-only | Distilled version of BERT | 91.2 (CoNLL-2003) | Faster inference, reduced parameters |
|---|---|---|---|---|
| **ALBERT** [15] | Encoder-only | BooksCorpus + English Wikipedia | 91.7 (CoNLL-2003) | Parameter sharing, memory efficient |
| **DeBERTa** [16] | Encoder-only | Enhanced Wikipedia + BookCorpus | 93.5 (CoNLL-2003) | Disentangled attention mechanism |

The performance of the advanced transformers is varied according to the architecture of the models and the text-dataset size. Therefore, to provide a good model in NER task, specifically, for Central-Kurdish dialectic text, a new text-dataset and modified architecture model are needed. Note, the Kurdish Language has several dialectics, and the text of each dialectic has its own syntaxes and even they have different semantic concept. Due to this reason, this study initially focused only on the Central-Kurdish dialectic. Toward this purpose, this study proposes a new fine-tuned architecture of RaBERTa and thus trained the tuned model via new Central-Kurdish dialectic corpus for the Kurdish NER task.

## 3. Methodology

The proposed approach started by collecting and creating Central-Kurdish dialectic corpus from different sources. The text-sources are online articles, news websites, and government publications to ensure a rich and varied representation. Note, the text size is around 1472 sentences. Thereafter, a set of preprocessing steps are applied such as: normalization, standardization, sentence segmentation, and tokenization. The overall proposed approach is demonstrated in Figure 1.

### 3.1 Preprocessing and Text Collection

**Normalizing** the Central-Kurdish dialectic text is an essential pretreatment step in text processing, particularly when working with informal text data. For example, normalization transforms the text into Unicode UTF-8 format. By utilizing this method, it can manage a multitude of character encodings, standardizing the text according to its protocols, and unifying characters. Specifically, certain terms that were written on an Arabic keyboard were converted to their Kurdish equivalents using the keyboard. For instance, the following substitution was made:

[(دوو ملیۆن و حەفت سەد هەزار) replaced by (دوومليۆن و حەفت سەد هەزار)]

This is followed by **standardizing** the character encoding, inserting spaces around punctuation marks, numerals, dates, and URLs, and eliminating the zero punctuation marks, numerals, dates, and URLs. Further, eliminating the Zero Width Non-Joiner (ZWNJ), i.e. the corpus undergoes conversion to the ZWNJ. For instance, the following substitution was made:

[(لە شاری کۆیە نزیک ٦٧٠٠ فەرمانبەر تۆمارکراوە) replaced by ( فەرمانبەر تۆمارکراوە 6700 لەشاری کۆیە نزیک )]

Note, the AsoSoft[2] tool [17] is used to standardize and normalize text in the Sorani dialect for consistency.

---

[2] https://github.com/AsoSoft/AsoSoft-Library-py

Further, for more uniformity of the text corpus, removing redundant characters and corrected typos are also applied. The next step of the text corpus creation is to apply text **segmentation**; this is to segment the text into sentences. Figure 2 shows the text segmentation into a set of sentences.

For **tokenizing** sentences, the Byte-Pair Encoding (BPE) is employed to handle sub-words and compound structures effectively, so that it will be aligned with RoBERTa's approach. Consequently, the **annotation** or labeling for the tokens are applied. To the best of our knowledge, with this study, the first annotated NER corpus specifically for the Sorani dialect of the Kurdish language is constructed. The corpus is created by collecting and annotating 1,500 sentences from various sources. The corpus employed a comprehensive set of tags for various named entities, as they are listed in Table II.

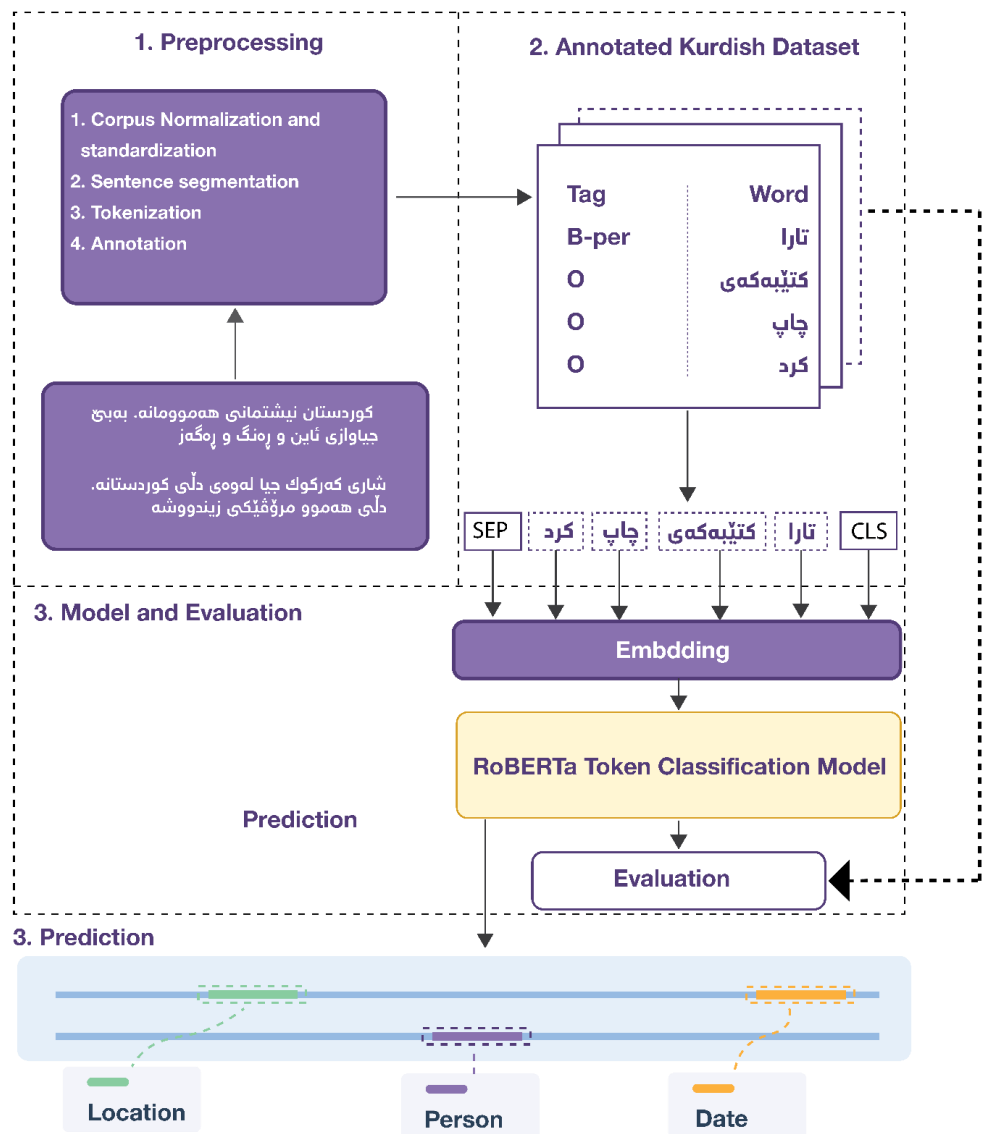

Figure 1. General block diagram of the proposed approach.

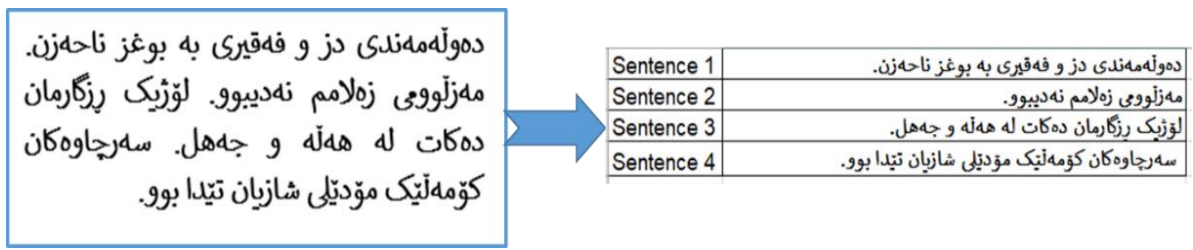

**Figure 2.** A snapshot of the text segmentation into sentences.

**Table II.** The list of the named entity with corresponding tags.

| Tag | Description |
| --- | --- |
| **B**-art | Begin Artifact/entity |
| **B**-ani | Begin Animal's name |
| **B**-bird | Begin Bird's name |
| **B**-dat | Begin Date |
| **B**-eve | Begin Event |
| **B**-fruit | Begin Fruit's name |
| **B**-geo | Begin Geographic entity |
| **B**-gpe | Begin Geopolitical entity |
| **B**-nat | Begin Natural phenomenon |
| **B**-org | Begin Organization name |
| **B**-per | Begin Person's name |
| **B**-tim | Begin Time expression |
| **B**-vine | Begin Vine's name |
| **B**-Money | Begin Monetary value |
| **B**-num | Begin Number |
| O | Non-Named Entity |
| I-gpe | Included Geopolitical entity |
| I-eve | Included Event |
| I-dat | Included Date |
| E-org | Ended Organization name |
| I-num | Included Number |
| I-per | Included Person name |

## 3.2 Text Annotation

Now, the list of tags is defined accordingly as well as the tokenized text is ready to be annotated. A professional staff of Central-Kurdish dialectic language persons are engaged to do the annotating tokens, manually via professors from the both University of Kurdistan Hewler, and

University of Sulaimani. Table III shows a summarized example of the annotated corpus which includes two sentences with corresponding tokens and their tags[3].

Table III. The list of the annotated tokens and their tags within two sentences.

| Sentence | Word | Tag |
|---|---|---|
| Sentence 1 | تارا | B-per |
| Sentence 1 | کتێبەکەی | O |
| Sentence 1 | چاپ | O |
| Sentence 1 | کرد | O |
| Sentence 2 | مشک | B-ani |
| Sentence 2 | لە | O |
| Sentence 2 | پشیلە | B-ani |
| Sentence 2 | دەترسێت | O |

This annotated format illustrates that how the tokens in the corpus tags within sentences identify entities and non-entity components. The structured data laid a strong foundation for building and training NER models specifically tailored for the language of Central-Kurdish dialect. Furthermore, Table IV show metadata and statistics on the number of tokens, number of sentences, and number of unique tags.

TABLE IV The Metadata of the created Kurdish NER Dataset.

| Metadata | Count |
|---|---|
| **Number of Sentences** | 1472 |
| **Number of Unique Tags** | 42 |
| **Total Number of Tokens** | 9528 |

In addition to the metadata of the NER Kurdish dataset, one more experiment has been conducted to show how the utilized tokens regarding to their tags are frequently distributed, as it can be depicted in Table V.

TABLE V Tag Frequency Distribution in KNER Dataset.

| Name of Tages | Count |
|---|---|
| O | 8194 |
| B-gpe | 207 |
| B-per | 168 |
| B-num | 149 |
| B-org | 111 |
| I-org | 90 |
| B-dat | 74 |
| I-per | 67 |

---

[3] https://huggingface.co/datasets/abdulhade/Kurdish_NameEntityRecognition

| | |
|---|---|
| **B-nat** | 59 |
| **B-tim** | 52 |
| **B-art** | 44 |
| **B-ani** | 40 |
| **B-eve** | 38 |
| **I-num** | 27 |
| **B-geo** | 27 |
| **I-gpe** | 25 |
| **I-eve** | 22 |
| **I-dat** | 13 |
| **E-org** | 13 |
| **B-bird** | 10 |

### 3.3 Model Architecture

Creating learning models or re-training pre-defined models for low resource languages (i.e. small text corpus) incurs huge computations and faces new challenges. To this end, this study proposed a modified/tuned RoBERTa model for the Kurish-Sorani dialectic. Furthermore, this article retrained the original RoBERTa architecture, only adjusting the final layer for the multi-class Kurdish-NER task. The model's pre-trained weights were fine-tuned with a linear classification layer tailored to recognize the various named entities, as they are listed in table 2.

Originally, RoBERTa is an optimized version of Bidirectional Encoder Representations from Transformers (BERT) [18]. Furthermore, the standard layers of RoBERTa are multi-head attention, add & normalize, Feed-Forward Network (FFN), and again add & normalize layers. The original layers of the RoBERTa architecture are demonstrated in figure 3.

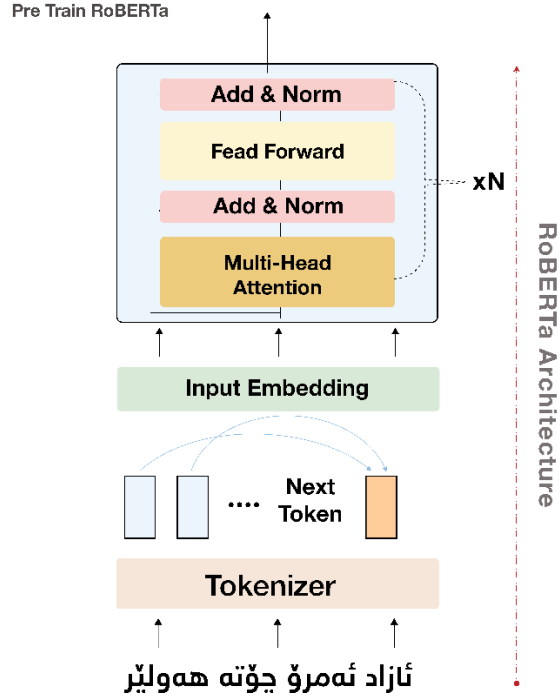

**Figure 3. The original architecture of the RoBERTa.**

Specifically for a single transformer layer in RoBERTa, the multi-head attention is described as $H_{attn}$, as the model output should be calculated, and it is expressed in equation (1):

$$H_{attn} = MultiHeadAttention(Q, K, V) \qquad (1)$$

Where the Q, K, and V are the query, Key and value metrics that derived from input *X*, the $H_{attn}$ is the output of the multi-head self-attention and it contains the context-aware representations of the input tokens after attention. Further, the Q represents the current token that should be compare with other tokens in the sentence, the K holds the information about all the token to compare against, finally the V carries the information that to be aggregated via the attention scores.

Thereafter, add and normalize layer is applied as follows:

$$H_{add1} = LayerNorm(X + H_{attn}) \qquad (2)$$

Where $H_{add1}$ is the combination of X input (the token embeddings or previous layer output) with the $H_{attn}$, this is a residual connection that helps with gradient flow during training. The *LayerNorm* denotes the normalization layer which ensuring that the output maintains a stable distribution for subsequent processing.

Consequently, the normalized embeddings through the **Feed-Forward Network** should be passed, as expressed in equation (3):

$$H_{ffn} = FFN(H_{add1}) \qquad (3)$$

Where the $H_{ffn}$ is refining the normalized embeddings by expands the dimensionality of $H_{add1}$ and reduces it back to the original dimension. This is applied via two linear transformations with a non-linear activation (typically **ReLU** or **GELU**).

Finally, Add and normalize layer is again applied to residual connections and normalize to produce the final output, as expressed in equation (4):

$$H = LayerNorm(H_{add1} + H_{ffn}) \quad (4)$$

Where the $H$ is the combination of $H_{add1}$ (i.e. output from the first Add & Normalize) with $H_{ffn}$ (i.e. output from the FFN). This is another **residual connection**. The $LayerNorm$ is to applying Layer Normalization to stabilize the output.

Each of these layers contributes to the model's ability to capture relationships and dependencies between words, making RoBERTa a powerful model for natural language understanding tasks.

However, the layers within RoBERTa model need further fin-tuning process, specifically for low-resources languages, specific domain (e.g. medical term), and extending for multilingual tasks. Further, when fine-tuning large pre-trained language models like RoBERTa, certain challenges arise including 1) **efficiency**: fine-tuning all parameters can be computationally expensive and memory intensive, 2) **overfitting**: for smaller downstream datasets, training all model parameters might lead to overfitting, and 3) **task-specific adaptation**: while frozen models work for general purposes, downstream tasks may require task-specific feature extraction. To do so, in this article, the modified or fine-tuned model of RoBARTa is proposed for Kurdish NER, as shown in figure 4.

The modified RoBERTa model is the modification of the standard transformer layer with zero-initialized attention. This modification is about an additional attention procedure into the standard transformer layer. First, the zero-initialized attention (it is also known Adapter) adds a new attention procedure which is initialized with **zero weights**, allowing the layer to adapt without affecting the original model's performance, as expressed in equation (5).

$$H_{adapter} = ZeroInitAttention(Q.K.V) \quad (5)$$

Where $H_{adapter}$ is a learnable attention output that starts as zero which effectively allows the model to slowly learn an additional layer of attention while keeping the original multi-head attention accordingly. In addition to that, the main advantage of zero-initialized attention is to 1) integrating zero weight-values into the original multi-head self-attention so that it doesn't affect the model's output and only the parameters of the adapter are trained, allowing task-specific representations to emerge,

Thereafter, the modified model combines the outputs ($H_{combined}$) of the original multi-head attention ($H_{attn}$) and the new zero-initialized attention ($H_{adapter}$). This allows the new attention procedure to enhance the representation over time while preserving the standard transformer's initial behavior, as expressed in equation (6):

$$H_{combined} = H_{attn} + H_{adapter} \quad (6)$$

The layer of add & normalize is applied on the combined output ($LayerNorm$) to: 1) keep the residual connection of the input $X$ and 2) ensure the stability and smooth gradient flow via the normalization ($LayerNorm$), as expressed in equation (7):

$$H_{add1} = LayerNorm(X + LayerNorm) \qquad (7)$$

This is followed by applying the two-layer feed-forward ($FFN$), same as standard transformer layer, as express in equation (8):

$$H_{ffn} = FFN(H_{add1}) \qquad (8)$$

In the last stage, both $H_{add1}$ and $H_{ffn}$ are combined via applying add & normalize layer for additional residual connection and to produce the final output by the normalization process, as expressed in equation (9):

$$H = LayerNorm(H_{add1} + H_{ffn}) \qquad (9)$$

The modified model is also fine-tuned with zero-initialized attention, and hence the fine-tuned introduces a focused training approach. To this end, in this study the original multi-head attention weights ($H_{attn}$) and Feed-Forward Network (FFN) weights are kept **frozen** during fine-tuning. This helps the modified model to provide pure model features. This means that only the ($H_{adapter}$) is trained and would be updated during fine-tuning procedure. Further, such a mechanism allows to learn the model to a new task by utilizing only the weights of the adapter, while preserving the general knowledge stored in the pre-trained parameters. However, for the initial state, the zero-initialized attention ignores the ($H_{adapter}$), thus the $H_{combined}$ contains only the $H_{attn}$, as it is expressed in equation (10):

$$H_{adapter} = 0$$

$$H_{combined} = H_{attn} \qquad (10)$$

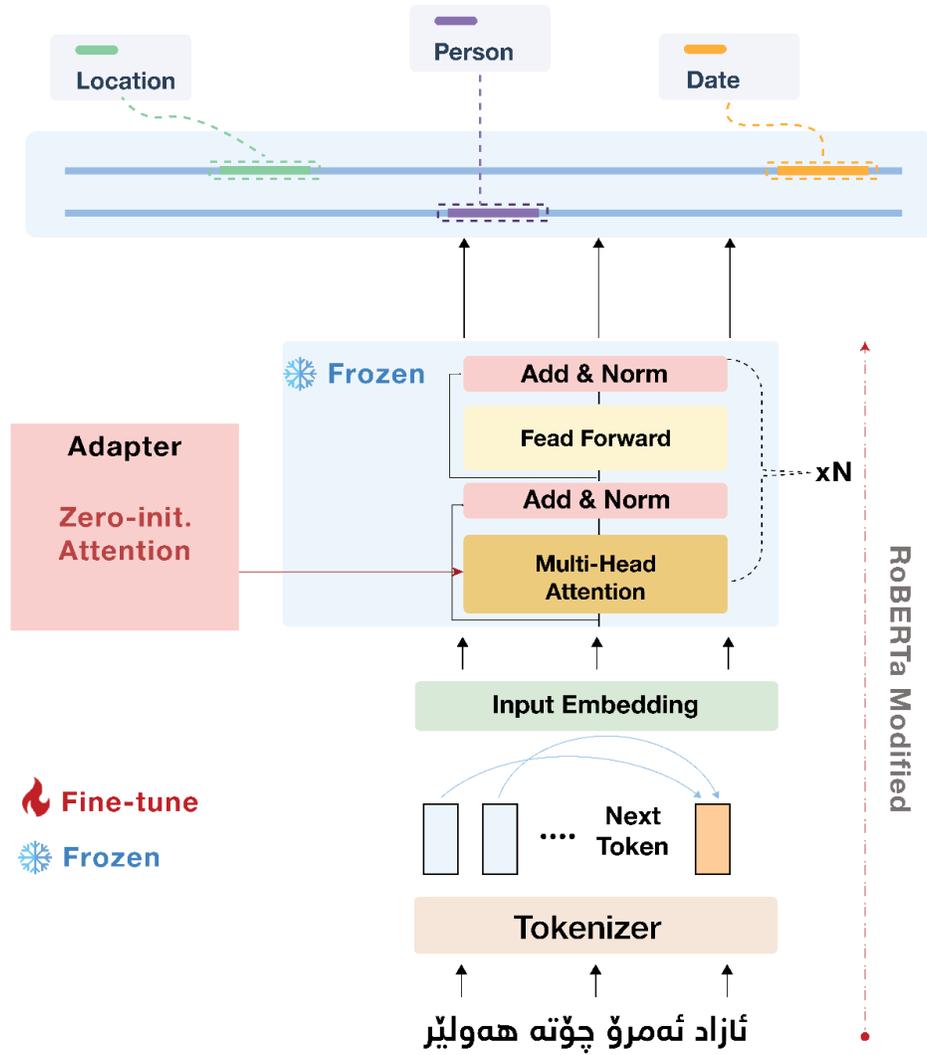

**Figure 4. The fin-tuned RoBERTa model for Kurdish NER.**

The final stage of fine-tuned output, after the training procedure, the adapter learns as a task-specific attention, thus the final layer output will be expressed in equation (11):

$$H^{fine-tuned} = LayerNorm(H_{atten} + H_{adapter}^{fine-tuned} + H_{ffn}) \qquad (11)$$

Where the $H_{adapter}^{fine-tuned} \neq 0$

## 4. Experiments and Evaluation Results

This section evaluates the performance of the proposed Kurdish Named Entity Recognition (KNER) model and compares it with baseline methods using different tokenization strategies such as Byte Pair Encoding (BPE), word-level, and sentence-piece. Basically, the model is fine-tuned using training data, updating hyperparameters, optimizer, and loss function. During the training model, the annotated Kurdish corpus or the dataset is divided into 70% for training subset, 15% for validation subset, and 15% for test subset. The hyperparameters of the model are configured when the learning rate is tuned to $3^{e-5}$, the batch size is set to 16, and repeated training process via 10 epochs. Further, the cross-entropy loss to optimize classification performance is utilized.

Subsequently, several experiments are conducted using different tokenization methods (including BPE, word-level, and sentence-piece) and using different models (the proposed/modified RoBERTa, original RoBERTa with Zero-shot, RoBERTa with LSTM, RoBERTa with SVM, and RoBERTa with MLP. The precision, recall, F1-score, and accuracy are main evaluation metrics to show the performance of the tokenization methods and trained models, as they are illustrated in table 3.

Table VI. The performance of tokenization methods and trained models.

| Model | Tokenization Method | Precision (%) | Recall (%) | F1 Score (%) | Accuracy (%) |
|---|---|---|---|---|---|
| RoBERTa (Proposed) | BPE | 92.4 | 91.8 | 92.1 | 91.5 |
| **RoBERTa (Proposed)** | **Sentence-piece** | **93.2** | **92.6** | **92.9** | **92.3** |
| RoBERTa | Word-level | 88.3 | 87.5 | 87.9 | 87.0 |
| RoBERTa Zero-shot | BPE | 79.5 | 78.3 | 78.9 | 78.0 |
| RoBERTa Zero-shot | Sentence-piece | 80.6 | 79.7 | 80.1 | 79.5 |
| Hybrid RoBERTa-LSTM | Word-level | 83.7 | 82.5 | 83.1 | 82.0 |
| Hybrid RoBERTa-LSTM | BPE | 84.2 | 83.1 | 83.6 | 83.0 |
| Hybrid RoBERTa-LSTM | Sentence-piece | 85.3 | 84.5 | 84.9 | 84.3 |
| Hybrid RoBERTa-SVM | Word-level | 77.2 | 76.1 | 76.6 | 75.5 |
| Hybrid RoBERTa-SVM | BPE | 78.4 | 77.3 | 77.8 | 77.0 |
| Hybrid RoBERTa-SVM | Sentence-piece | 79.6 | 78.8 | 79.2 | 78.5 |
| Hybrid RoBERTa-MLP | BPE | 80.3 | 79.9 | 80.1 | 79.5 |
| Hybrid RoBERTa-MLP | Sentence-piece | 81.5 | 81.0 | 81.3 | 80.8 |

As the sentence-piece tokenization method across all the trained models achieved the highest F1-scores and accuracy due to its superior handling of subword-level segmentation, particularly effective for Kurdish language's rich morphological structure. In addition, the proposed fine-tuned RoBERTa with sentence-piece tokenization achieved the best overall performance, setting new benchmarks with an F1 score of 92.9% and an accuracy of 92.3%. While hybrid RoBERTa-LSTM and hybrid RoBERTa-MLP demonstrated solid improvements when using sentence-piece tokenization, achieving F1 scores of 84.9% and 81.3%, respectively. On the other hand, the zero-shot classification is the weakest performer overall, even with sentence-piece tokenization, highlighting the importance of fine-tuning for specific tasks.

Finally, to illustrate the performance gap between the models, table 5 shows F1 score improvements achieved by the proposed model over baseline models and it highlights the comparative performance improvements. As the results of fine-tuned models outperform the zero-shot by a 12.8% improvement in F1 score, this is showcasing the importance of domain-specific training. Furthermore, despite their simple architecture, the LSTM and SVM achieved notable F1 scores but lagged transformer-based models, emphasizing their limited capacity to capture complex contextual relationships. The hybrid RoBERTa-MLP model performed well with sentence-piece tokenization but still fell short of the proposed RoBERTa model due to the superior architecture of the latter.

Table VII. A comparative performance gain (measured by f1-score) of the proposed models.

| Model Comparison | Tokenization Method | F1 Score Improvement Over Baseline (%) |
|---|---|---|
| Proposed RoBERTa vs. Zero-shot | Sentence-piece | +12.8% |
| Proposed RoBERTa vs. LSTM | Sentence-piece | +8.0% |
| Proposed RoBERTa vs. SVM | Sentence-piece | +13.7% |
| Proposed RoBERTa vs. MLP | Sentence-piece | +11.6% |

## 5. Conclusion

The comparative performance analysis reported in this work demonstrates that the proposed RoBERTa model, fine-tuned with sentence-piece tokenization, establishes a new benchmark for Kurdish NER tasks. The results reinforce the significance of advanced transformer architectures, effective tokenization strategies, and task-specific fine-tuning in improving low-resource language processing. Future work will focus on expanding datasets and incorporating additional dialects to further enhance model robustness. This study also demonstrates that with strategic data collection and model adjustments, transformer models can be adapted for low-resourced languages, paving the way for further NLP advancements in these domains. The promised solution of zero-initialized attention is due to task adaptability without re-training the features of the RoBERTa, requires fine-tuning only a small subset of parameters (adapter weights), reducing computational overhead, maintaining robustness, and providing better performance. However, the limited high quality of annotated data, the variety of Kurdish language dialectics, and complex morphology of Kurish words are still the main challenges for the future studies that needs to be addressed.